\documentclass{article}

    \PassOptionsToPackage{numbers, compress}{natbib}

\usepackage[final]{neurips_2022}

\usepackage[utf8]{inputenc} %
\usepackage[T1]{fontenc}    %
\usepackage{hyperref}       %
\usepackage{url}            %
\usepackage{booktabs}       %
\usepackage{amsmath}        %
\usepackage{amsfonts}       %
\usepackage{microtype}      %
\usepackage{xcolor}         %
\usepackage{graphicx}
\usepackage[capitalise,nameinlink]{cleveref}
\usepackage{paralist}
\usepackage{caption}        %
\usepackage{subcaption}     %
\usepackage{scalerel}       %
\usepackage{tikz}           %
\usepackage{bm}             %
\usepackage[export]{adjustbox} %
\usepackage{multirow}       %
\usepackage{tabularx}       %
\usepackage{wrapfig}
\usepackage{pifont}
\usepackage{physics}        %
\usepackage{rotating}
\usepackage{algorithmic}
\usepackage{algorithm}

\crefname{section}{§}{§§}
\Crefname{section}{§}{§§}
\crefformat{equation}{(#2#1#3)}
\renewcommand{\O}{\mathcal{O}} %
\newcommand{\M}{\mathcal{M}} %
\newcommand{\V}{\mathcal{V}} %
\renewcommand{\P}{\mathcal{P}} %
\renewcommand{\L}{\mathcal{L}} %

\title{Learning to Infer 3D Shape Programs\\ with Differentiable Renderer}

\author{%
  Yichao Liang\thanks{Work done in 2020 during an internship with Bob Fisher, The University of Edinburgh.} \\
  Department of Computer Science\\
  University of Oxford\\
  \texttt{ycliang6@gmail.com} \\
}

\begin{document}

\maketitle

\begin{abstract}
Given everyday artifacts, such as tables and chairs, humans recognize high-level regularities within them, such as the symmetries of a table, the repetition of its legs, 
while possessing low-level priors of their geometries, e.g., surfaces are smooth and edges are sharp.
This kind of knowledge constitutes an important part of human perceptual understanding and reasoning. 
Representations of and how to reason in such knowledge, and the acquisition thereof, are still open questions in artificial intelligence (AI) and cognitive science.
Building on the previous proposal of the \emph{3D shape programs} representation alone with the accompanying neural generator and executor from \citet{tian2019learning},
we propose an analytical yet differentiable executor that is more faithful and controllable in interpreting shape programs (particularly in extrapolation) and more sample efficient (requires no training).
These facilitate the generator's learning when ground truth programs are not available, and should be especially useful when new shape-program components are enrolled either by human designers or---in the context of library learning---algorithms themselves.
Preliminary experiments on using it for adaptation illustrate the aforesaid advantages of the proposed module, 
encouraging similar methods being explored in building machines that learn to reason with the kind of knowledge described above, and even learn this knowledge itself.
\end{abstract}

\section{Introduction}
We study the inverse problem of inferring holistic representations from 3D shapes.
When given a 3D shape object (e.g., a chair), we are interested in inducing a graphics program defined in a Domain-Specific Language (DSL) that captures, in particular, the high-level regularities of the 3D object, as well as the low-level shape priors.
Using a program-like latent representation brings forth notably the following three advantages.
\begin{inparaenum}
    \item Given images of real-life objects, such as tables, office chairs, humans immediately recognize the high-level concept embodied in the objects such as reflective symmetries of tables, radial symmetries of the legs of office chairs, etc. 
    This kind of knowledge has been shown to constitute an important aspect of human visual recognition and reasoning \cite{koffka2013principles, dilks2011mirror}, and is naturally expressed by programming language constructs such as the \texttt{For loop} statement.
    \item By using shape primitives as the terminals in a DSL grammar, we are able to capture the shape priors such as sharp edges and smooth surfaces that the other shape representations such as voxels, point clouds, meshes find hard to capture.
    \item Finally, such high-level, holistic object representations can facilitate low-level manipulation tasks such as inpainting and extrapolation.
\end{inparaenum}

\begin{figure*}[!th]
    \centering
    \includegraphics[width=\linewidth]{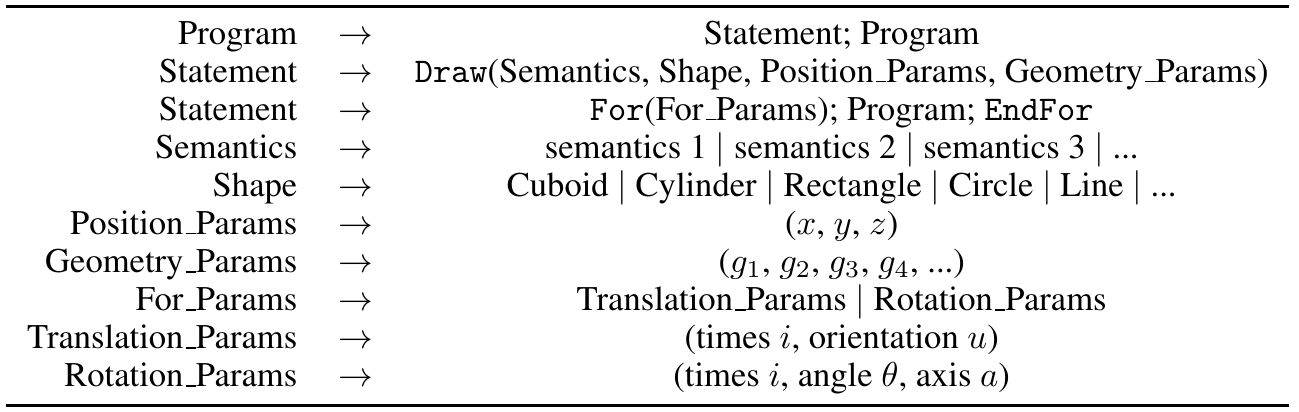}
    \caption{The DSL for 3D shape programs, copied from \cite{tian2019learning}.}
    \label{fig:dsl}
\end{figure*}

In recent years, there is a particular surge of interest in modelling perceptual data using graphics programs. 
These include focuses on 2D images \cite{Mao_2019_ICCV, li2020perspective}, 3D shapes \cite{tulsiani2017learning, tian2019learning}, or even scenes and short videos \cite{li2020multi, liu2018learning}.
Most inspiring to us, \citet{tian2019learning} propose \emph{3D shape programs}---as defined by a DSL (\cref{fig:dsl})---as a useful representation for modeling 3D shapes.
The DSL is accompanied by a pair of neural generator-executor (a.k.a. encoder-decoder) for inferring and interpreting shape programs respectively.
In particular, the neural program executor is motivated by the need for finetuning the generator in the absence of ground truth programs 
and by the understanding that certain high-level program command, e.g., the \texttt{For loop} statement would make any analytical renderer \emph{non-differentiable} \cite{tian2019learning}.
In our work, we argue that the \texttt{For loop} statement and the \emph{shape programs} in general can indeed be executed by a non-neural yet still differentiable renderer that we contribute.

The analytical renderer is not only more precise and requires fewer parameters, which could be important in finetuning;
but also make enrollment of new shape primitives much more straightforward, which could be particularly useful in the context of library learning as neural alternatives would likely require additional data and training for the new shape program components.

In summary, our contributions are mainly two-fold:
\begin{enumerate}
    \item we contribute a differentiable rendering method for \emph{shape programs}, which can lead to more effective adaptation to new classes of objects.    
    \item we suggest several methods of training the program generator when such a renderer is available, and show preliminary results when one of these objectives is employed.
\end{enumerate}

\section{Related work} %
Many recent works attempt to capture program-like perceptual abstraction of visual data.
For 2D images, \citet{Mao_2019_ICCV} demonstrate the ability to capture the regularities (lattice, circular pattern) of repeated objects' position in the scene captured from a \emph{fronto-parallel} plane with respect to the object plane, with a search-based algorithm. 
\citet{li2020perspective} exploit the regularities among the objects, generalizing the previous work to allow arbitrary viewpoints, 
by explicitly estimating the camera pose as a part of the program synthesize process in a hybrid of search and gradient-based algorithm.
\citet{liu2018learning} model regularly placed primitive objects in images with not only centroids but the shape primitives themselves, such as cubes, spheres, and cylinders, alone their position and rotation in the scene. 
On the front of 3D objects, \citet{tulsiani2017learning} model 3D voxel representation of more complex objects, e.g., chairs, tables, with cuboids but without using high-level program statements, e.g., the \texttt{For loops}, which results in difficulties in capturing regularities in objects and the semantics of primitives.
\citet{sharma2019neural} use programs in a language of Constructive Solid Geometry (CSG) (which allows arithmetic operations on shapes) to enable modeling of more complex shapes.
\citet{tian2019learning} performs program synthesis on similarly complex objects in canonical view frames, directly from 3D voxel, using purely a neural approach.
Their system leverages the common bilateral and/or radial symmetry of artifacts to achieve higher accuracy than \citet{tulsiani2017learning}. 
To allow self-supervised learning without program-level annotation, they propose another neural module, the program executor, for rendering programs in their DSL.

Their system is trained through a biphasic procedure: 
\begin{inparaenum}
    \item train each of the two modules on hand designed, synthetic data, in a supervised learning setting;
    \item train the program generator with gradient signals backpropogated through the program executor computed on reconstruction loss with respect to real data, e.g., from ShapeNet.
\end{inparaenum}
The latter is named as \emph{Guided Adaptation} in the sense that the neural program executor guides the generator to adapt to out-of-distribution object classes.

\begin{figure*}[!th]
  \begin{subfigure}[b]{0.25\linewidth}
  \centering
  \includegraphics[width=.8\linewidth]{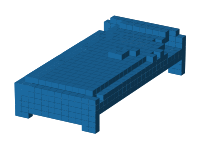}
  \caption{\label{fig:target1}}
  \end{subfigure}%
  \begin{subfigure}[b]{0.25\linewidth}
  \centering
	\includegraphics[width=.8\linewidth]{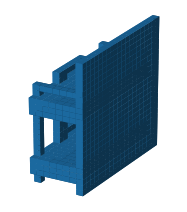}
  \caption{\label{fig:target2}}
  \end{subfigure}%
  \begin{subfigure}[b]{0.25\linewidth}
  \centering
	\includegraphics[width=.8\linewidth]{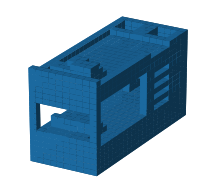}
  \caption{\label{fig:target3}}
  \end{subfigure}%
  \begin{subfigure}[b]{0.25\linewidth}
  \centering
	\includegraphics[width=.8\linewidth]{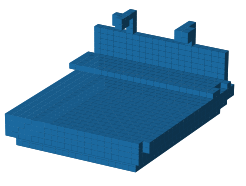}
  \caption{\label{fig:target4}}
  \end{subfigure}
  
  \begin{subfigure}[b]{0.25\linewidth}
  \centering
  \includegraphics[width=.8\linewidth]{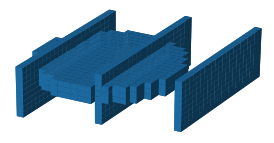}
  \caption{\label{fig:recon1}}
  \end{subfigure}%
  \begin{subfigure}[b]{0.25\linewidth}
  \centering
	\includegraphics[width=.8\linewidth]{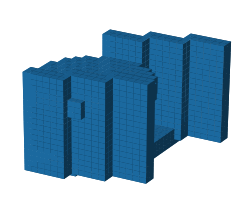}
  \caption{\label{fig:recon2}}
  \end{subfigure}%
  \begin{subfigure}[b]{0.25\linewidth}
  \centering
	\includegraphics[width=.8\linewidth]{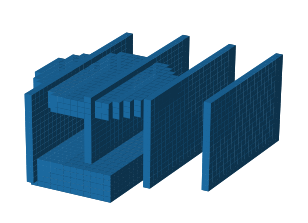}
  \caption{\label{fig:recon3}}
  \end{subfigure}%
  \begin{subfigure}[b]{0.25\linewidth}
  \centering
	\includegraphics[width=.8\linewidth]{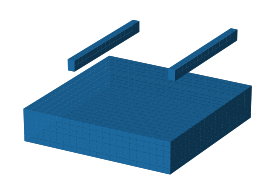}
  \caption{\label{fig:recon4}}
  \end{subfigure}

  \caption{Representative reconstruction samples after ``guided adaptation" from \citet{tian2019learning} from a random class of objects---beds. The first row shows the input shapes in from ShapeNet a voxel representation; the second row depicts the shape rendered from the predicted shape programs.}
  \label{fig:guided_result}
\end{figure*}

We note that the \emph{Guided-Adaptation} towards new subclasses of objects outside the synthetic ones, i.e., subclass are not seen during the training with synthetic data e.g., beds, benches, did not lead to ideal results, as demonstrated in \cref{fig:guided_result}. Many reconstructions are neither faithful nor consistent.
We hypothesize that a part of this side effect is due to the approximation nature of the program executor.
With this and its relatively poor sample efficiency, we regard our method as the first step towards more efficient and effective adaptation to new object classes and possible shape-program library learning.

\section{Problem Statement}
We review the problem of \emph{Shape Primitive Decomposition} and \emph{Shape Program Decomposition}.
Let $\O$ denote a 3D object, e.g., the 3D model of a chair, and $\M, \V, \P$ denote the mesh, voxel, and point cloud representation of $\O$, respectively.

\paragraph{Shape Primitive Decomposition.} Use $[n]$ to represent the set of integers $\{1,\dots,n\}$. 
Inspired by \citet{tulsiani2017learning} and \citet{zou20173d}, we formulate the problem of shape primitive decomposition as: 
given a shape object $\O$ and a set of $N$ shape primitives $\{P_n\}$ with $n\in[N]$, 
find a collection of $M$ primitives such that their affine transformed version $[\bar{P}_m]$ has a small discrepancy when placed together with respect to the target shape, i.e., $\L(\bigcup_m\bar{P}_m, \O)$ is small.
Note that this objective can in principle computed between any combination of primitive, target representations (such as voxel, mesh, etc).

In more detail, each transformed primitive $\bar{P}_m$ is specified by a 4-tuple $(P_m^i, P_m^z, P_m^t, P_m^q)$, where $P_m^i$ identifies its type, e.g., a cuboid, or an ellipsoid, etc;
$P_m^z$ its size, e.g., in case $P_m$ is a cuboid, its height, width, depth;
$P_m^t=(t_m^x, t_m^y, t_m^z)$ its center coordinate in the target frame, and
$P_m^q=(q_m^x, q_m^y, q_m^z)$ its rotation in Euler angle and axis form. 

\paragraph{Shape Program Decomposition.} 
Shape programs can be viewed as an extension to shape primitives that are able to more holistically captures the structures of shapes by leveraging control flow statements from imperative programming languages.  

In this context, the problem can be framed as, when given a shape object $\O$, infer a shape program $\rho$
such that the shape rendered from it has small discrepancy with respect to the target object $\L(R(\rho), \O)$---using $R(\cdot)$ to denote the interpreter for the language.
This is similar in flavor to the objective in shape primitive decomposition.

\section{Approach}
Building on the neural program generator (a.k.a., encoder) from \citet{tian2019learning} (summarized in \cref{sec:prog_gen}), 
we describe in detail the analytical, differentiable renderer that we are contributing (\cref{sec:diff_rend}) 
as well as two concrete ways of finetuning the program generator in a self-supervised fashion. 
When used in tandem with our renderer, we refer to them as Direct Adaptation (DA).
The renderer we present here is more accurate, especially in extrapolation, and does not require any training compared to neural alternatives. 
Consequently, this should lead to a more preferable finetuning result and would be particularly convenient when additional shape-program components are enrolled by either designers or during library learning, as a neural renderer would require additional training.
 
\subsection{Program Generator}\label{sec:prog_gen}
At a high level, given the voxel representation $\V$ of a shape $\O$, the program generator $h(\cdot)$ predicts the corresponding shape program consisting of $K$ program blocks, each with $J$ steps, that attempts to capture the geometry of $\V$.
The inferred programs are defined by a DSL (\cref{fig:dsl}). 
We use exactly the same design as in \citet{tian2019learning} for the later experiments to compare the benefit of using different renderers.

\begin{figure*}
    \centering
    \includegraphics[width=\linewidth]{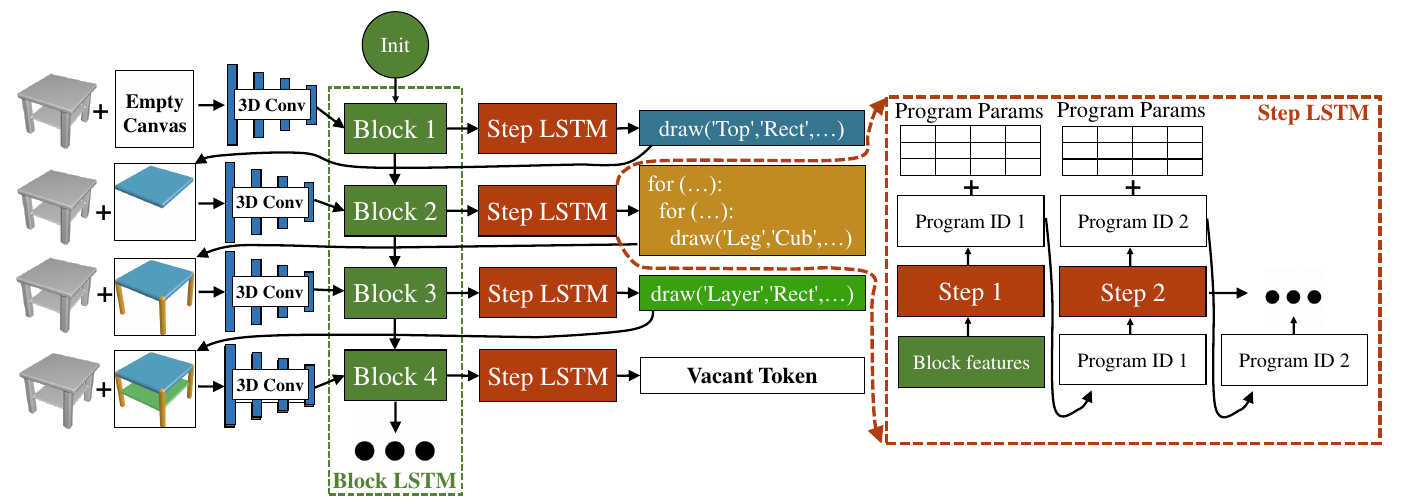}
    \caption{Schematics borrowed from \citet{tian2019learning} to illustrate the design of the shape program generator. Exactly the same design is used in our experiment to evaluate the renderer that we are proposing. The architecture is composed of three modules: a 3D convolution network, a Block, and a Step LSTM.}
    \label{fig:generator}
\end{figure*}

In more detail, as illustrated in \cref{fig:generator}, the generator processes an input voxel in the following three steps:
\begin{inparaenum}
\item a 3D convolution net takes the stack of the target voxel and the voxel-reconstruction-so-far, to generate a feature encoding, $h^k_{shape}$;
\item a LSTM (denoted as the block LSTM) uses $h^k_{shape}$ to compute a block-level feature $h^k_{block}$;
\item finally, another LSTM (the step LSTM) takes $h^k_{block}$ to produce a fixed, $J$-steps of program statements. 
At each step, the output consists of $G$ logits for the categorical distribution of the program statement (including one for the \texttt{empty} statement) and $L$ values for parameterizing the sampled statement, independent of the final statement chosen.
\end{inparaenum}

\subsection{Direct Adaptation with Differentiable Executor}\label{sec:diff_rend}
In the absence of program-level supervision, as it is usually the case, after learning on synthetic data, the generator can primarily only be trained with variations of the reconstruction tasks; 
compute a reconstruction loss of the rendering output with respect to any representation of the target object $\O$ and backpropagate the gradient to the generator for updates. 

To this end, one can either leverage a neural program executor---as in \cite{tian2019learning}---or an analytical, differentiable renderer, which is introduced in this section. 
To be specific, a differentiable renderer transforms a shape program predicted by the renderer, $\rho = [\rho_{kj}]$, 
to a graphical representation of the program such as a point cloud or voxel, 
which can be used in computing differentiable loss objectives, for finetuning the generator's parameters.

The key insight that enables analytical yet differentiable rendering in the presence of high-level statements such as the \texttt{For loop} statement is that 
each block of the generated program $\rho_m$ can be unrolled and transformed, via a differentiable operation, into a semantically equivalent, collection of shape primitives, $\{(P_n^i, P_n^z, P_n^t, P_n^q)\}$, as in the problem of \emph{shape primitive decomposition}.
For example, the \texttt{For loops} with shape primitive statement inside them, can be expended into a set of basic drawing statements with the renderings before and after the expansion correspond to the same final shape.

The procedure of removing higher-order statements is performed on a case-by-case basis as they are parameterized in distinct spaces.
As a concrete example, a \texttt{Translation For} statement $\rho$ is parameterized by $(\rho_1, \rho_2, \rho_3)$ representing the incremental shift on the $(x, y, z)$ axes of the objects inside the loop, 
while a \texttt{Rotation For} statement has only one meaningful parameter, $(\rho_1)$, indicating the number of times the inner drawing statement should be repeated with respect to the x-axis to complete a full rotation, 
e.g., $\rho_1=3$ denotes the inner statements are drawn 3 times with 120-degree rotation apart, and $\rho_1=5$ means 5 repetitions, 72-degree apart.

Different primitive statements $\rho$ also have distinct parameter spaces, which would need to be taken into consideration both when they are inside higher-order statements 
and when converting them from their own program parameter spaces $[\rho_l]$ to the shape primitive space, $(P^i, P^z, P^t, P^q)$.
To illustrate, a \texttt{line} drawing statement draws a cylindrical shape that has $(\rho_1, \rho_2, \rho_3), (\rho_4, \rho_5, \rho_6)$ to indicate its start, end coordinates respectively, and $(\rho_7)$ to specify its radius; 
While a \texttt{chair\_back} statement has $(\rho_1, \rho_2, \rho_3)$ representing the front-bottom-left position of a cuboid, $(\rho_4, \rho_5, \rho_6)$ specifying its height, width and depth, 
and $(\rho_7)$ representing its elevation angle. 
In total, 22 semantically distinct programs statements exist in the original DSL.

With this in mind, a shape program $\{\rho_{kj}\}$ is unrolled and replaced with the collection of semantically equivalent basic program statements, 
then the basic statements are transformed to the shape primitive space with $\{(P_n^i, P_n^z, P_n^t, P_n^q)\}$. 
In the end, these standardized shapes are either cuboids or cylinders for the current DSL. Up to this point, we have in our hand a rather abstract representation of the reconstructed shape; we haven't provided any method for ``displaying" them onto the screen.
Nevertheless, we can already derive a differentiable objective for learning, as we will describe in \cref{sec:mesh_w_primitives}.

We carry on to present a way of rendering the shapes primitives to a point cloud representation, exercising the \emph{reparameterization trick} in a similar spirit to \cite{tulsiani2017learning, kingma2013auto}.
We exploit the fact that sampling a point $p$ on a transformed primitive $\bar{P}$, is semantically equivalent to sampling a point $p'$ from a standard uniform distribution, $p' \sim \mathcal{U}[-1,1]^3$, then transforming it according to the shape's parameters $(P^z, P^t, P^q)$. 
Compared to sampling directly from the range defined by $\bar{P}_m$, this method has the crucial benefit of decoupling the outputting parameters from the stochastic sampling process, which enables backpropagation of the loss gradient to the shape's geometric parameters. As an example, to sample a point from the lateral surface of a cylinder $\bar{P}_c$ with parameters $(P_c^z, P_c^t, P_c^q)$ where $P_c^z=(s_r,s_h)$, 
we can first sample an angle value $\theta \sim \mathcal{U}[-\pi, \pi]$ and a height $h \sim \mathcal{U}[-\frac{1}{2},\frac{1}{2}]$. 
Then the corresponding point on $\bar{P}_c$ can be obtained by applying the rotation $P^t_q$ and translation $P^t_c$ to the point $p''=(s_h * h, \text{sin}(\theta), \text{cos}(\theta))$

To ensure uniform sampling coverage w.r.t. all primitive surfaces corresponding to an object, 
the number of points sampled on each shape primitive's surface is determined to be proportional to its surface area.

Note that besides the abstract primitive representation and the point cloud representation we presented here, other differentiable representations could also be devised and employed (e.g.,  voxel, mesh, hidden activations from another neural net).
Further, while the geometry parameters of a shape program (including position, rotation, etc) may receive gradients when trained with these renderers, 
the program indices itself (i.e., whether to use the \texttt{table\_top} program or \texttt{chair\_leg} program or a For loop statement) does not. 
But this is independent of whether higher-level statements exist in the DSL, and gradient estimators for discrete variables such as REINFORCE and NVIL \cite{mnih2014neural, schulman2015gradient} can be utilized.

\subsection{Learning Objectives}
With the reconstructed representations, self-supervised learning objectives can be defined on top of nearly all (target representation, reconstructed representations) combinations.
We elaborate on two such objectives, building on the two representations we discussed in \cref{sec:diff_rend} respectively.
Namely, we introduce a loss for when target point clouds are paired with abstract shape primitives reconstructions (\cref{sec:mesh_w_primitives}), and with point clouds reconstructions (\cref{sec:pc_w_pc}).
Point cloud representations for target shapes are commonly available; 
they could be easily obtained from a mesh representation which is the default representation in many 3D shape datasets, 
e.g., ShapeNet \cite{chang2015shapenet} or derived from the voxel representation that the program generator uses.

\subsubsection{Point Cloud with Point Cloud}\label{sec:pc_w_pc}
We start with a straightforward objective defined with respect to a target point cloud and a reconstructed point cloud.
With a pair of point clouds, Chamfer distance (CD) can be readily computed and leveraged for learning:
it is computed by taking the sum of the square distances for each point in one point cloud to its nearest point in the other point cloud.
Note that by the definition as above, CD is not symmetrical in its arguments and hence not a mathematical distance function.
Thus one may wish to utilize a combination of the forward and backward score in optimization.

\subsubsection{Point Cloud with Shape Primitives}\label{sec:mesh_w_primitives}
With the set of shape primitives $\{\bar{P}_m\}$ representing the generator prediction, 
the coverage loss of the target point cloud $\P$ and the union of the shape primitives is another criterion for adapting the generator to target shapes without program supervision, as inspired by \citet{tulsiani2017learning}. 
 
The objective suggests that the generator should be penalized if the union of the generated shapes $\bigcup_m \bar{P}_m$ does not cover the target point cloud $\P$.
A sufficient condition to ensure this is that all sample points from the target point cloud evaluate zero in the distance field of the generated shapes:
\begin{equation}
   \L(\{\bar{P}_m\},\P) = \mathbb{E}_{p\sim \P} \left[\|\mathcal{C}(p;\bigcup_m \bar{P}_m)\|_2 \right]
\end{equation}
Where we make use of the notion of distance field $\mathcal{C}(p; P):: \mathbb{R}^3 \to \mathbb{R}^+$ defined on shape primitive $P$, denoting the closest distance from a point $p$ to the surface of the shape primitives $P$.

When calculating the distance field of a set of shape primitives, 
it can be shown that the distance field w.r.t. the union of a set of primitives equals to the element-wise minimum among the distance fields of the individual primitives.
\begin{equation}
    C(p;\bigcup_m \bar{P}_m) = \text{min}_m \mathcal{C}(p;\bar{P}_m)
\end{equation}

As we only have cuboid and cylinder primitives in the current DSL, if suffice to define the distance field for them.
For cuboid, the distance field $\mathcal{C}_{cub}(p;P^z)$ of an origin-centered cuboid with $P^z= (s_x, s_y, s_z)$ can be defined as in as\footnote{let max$(0,x) \equiv x_+$.}:
\begin{equation}
    \mathcal{C}_{cub}(p;P^z)^2 = (|p_x| - s_x)^2_+ + (|p_y| - s_y)^2_+ + (|p_z| - s_z)^2_+
\end{equation}
as in \cite{tulsiani2017learning}.
By a similar token, the distance field $\mathcal{C}_{cyl}(p;P^z)$ of an origin-centered cylinder with $P^z=(s_x, s_y)$ corresponding to the height, radius of the cylinder can be computed by a simple algorithm (for details, see \cref{app:cyl_df}). 

With the method of computing the distance field of primitives in their canonical frame, the distance field at a point $p$ of a transformed primitive $\bar{P}$ with transformation parameters $(P^t, P^q)$ can be calculated by evaluating the result of applying the inverse transformation of $P^t, P^q$ to the point $p$ in $\mathcal{C}(\cdot~;P)$. 
We thereby obtain a way to compute the distance field of any primitive in any pose:

\begin{equation}
    \mathcal{C}( \cdot ;P_m) =
    \begin{cases}
      \mathcal{C}_{cub}( \cdot ; P_m^z), & \text{if}\ P_m^i=\text{cuboid} \\
      \mathcal{C}_{cyl}( \cdot ; P_m^z), & \text{otherwise}
    \end{cases}
\end{equation}

\section{Experiments}

\begin{figure*}[!ht]
  \begin{subfigure}[b]{.5\linewidth}
  \centering
  \includegraphics[width=.9\linewidth]{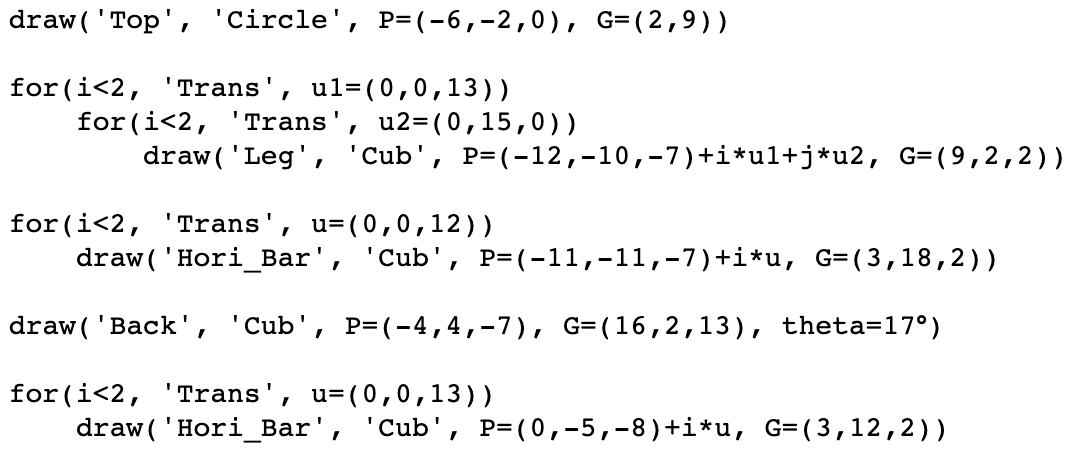}
  \caption{\label{fig:pg1}}
  \end{subfigure}%
    \begin{subfigure}[b]{.5\linewidth}
  \centering
  \includegraphics[width=.9\linewidth]{figures/pg1.png}
  \caption{\label{fig:pg2}}
  \end{subfigure}%
  
  \begin{subfigure}[b]{0.25\linewidth}
  \centering
  \includegraphics[width=.6\linewidth]{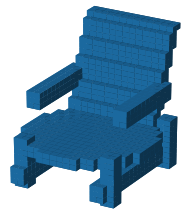}
  \caption{\label{fig:v1}}
  \end{subfigure}%
  \begin{subfigure}[b]{0.25\linewidth}
  \centering
	\includegraphics[width=0.5\linewidth]{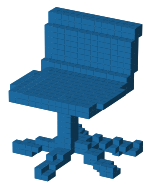}
  \caption{\label{fig:v2}}
  \end{subfigure}%
  \begin{subfigure}[b]{0.25\linewidth}
  \centering
	\includegraphics[width=.6\linewidth]{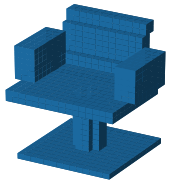}
  \caption{\label{fig:v3}}
  \end{subfigure}%
  \begin{subfigure}[b]{0.25\linewidth}
  \centering
	\includegraphics[width=.6\linewidth]{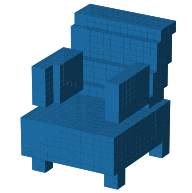}
  \caption{\label{fig:v4}}
  \end{subfigure}
  
  \begin{subfigure}[b]{0.25\linewidth}
  \centering
  \includegraphics[width=.6\linewidth]{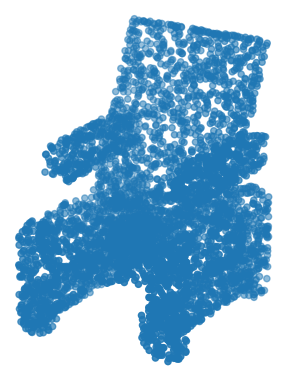}
  \caption{\label{fig:ren1}}
  \end{subfigure}%
  \begin{subfigure}[b]{0.25\linewidth}
  \centering
	\includegraphics[width=.6\linewidth]{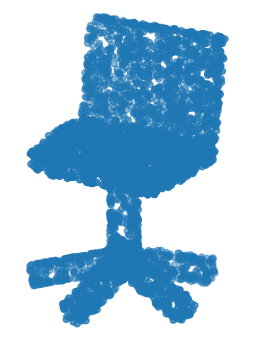}
  \caption{\label{fig:ren2}}
  \end{subfigure}%
  \begin{subfigure}[b]{0.25\linewidth}
  \centering
	\includegraphics[width=.6\linewidth]{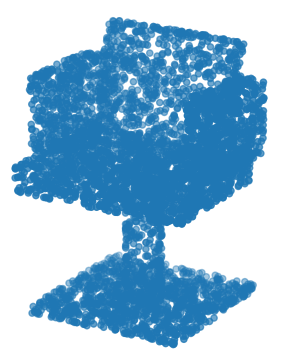}
  \caption{\label{fig:ren3}}
  \end{subfigure}%
  \begin{subfigure}[b]{0.25\linewidth}
  \centering
	\includegraphics[width=.6\linewidth]{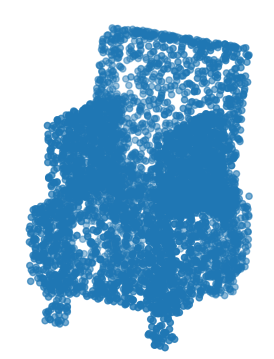}
  \caption{\label{fig:ren4}}
  \end{subfigure}

  \caption{The first row shows the shape program corresponding to the first 2 chairs below; the second row shows the rendered shapes from the non-differentiable render; and the third row shows the outputs of our differentiable renderer in the form of sampled surface points, which can be used to train the program generator model.}
  \label{fig:rendering_result}
\end{figure*}

We experiment with the accuracy (\cref{sec:exp_renderer}) and utility (\cref{sec:exp_finetuning}) of our differentiable renderer. 
We show preliminary results of direct adaptation that uses the point cloud representation and the chamfer distance objective.
We run experiment on a subset of ShapeNetCore v2. 
Each object $\O$ has a mesh representation $\M$ and \textit{binvox}\footnote{https://www.patrickmin.com/binvox/} is used to convert $\M$ to a voxel representation $\V$ of dimension 32x32x32 as the input for the neural program generator.

\subsection{Differentiable Rendering}\label{sec:exp_renderer}
Figure \ref{fig:rendering_result} shows sampled non-differentiable vs differentiable renderings, and their corresponding program outputs. 
The output of our renderer is illustrated by 5,000 sample points on the surface of the shape primitives.
Qualitatively, the differentiable renderer is able to accurately interpret different types of complex program statements, which is critical to the generator's successful adaptation to new classes of shapes.
And we hope to perform more quantitative evaluations in future works.

\subsection{Results of Direct Adaptation}\label{sec:exp_finetuning}
By sampling 5,000 points on the reconstructed shape primitives' surfaces and the mesh surface respective, we finetune the generator on the Chamfer distance loss as defined in \cref{sec:pc_w_pc}. 
\cref{tab:performance} shows that our model outperforms the Guided Adaptation model by more than. 50\% as measured by Chamfer distance on the held out set.
We leave further qualitative and quantitative evaluations for future work.

\begin{table}[!ht]
    \caption{The performance of the Shape Program Generator after being fine-tuned with different adaptation methods. The first row shows the model without any adaptation.}
    \vskip 0.15in
    \centering
    \begin{tabular}{|l|c|c|}
    \toprule
    Models & CD bench & CD cabinet\\ 
    \midrule
    Shape Programs & 0.0135 &  0.0104\\
    Shape Programs w. GA & 0.0078 & 0.0092\\ 
    Shape Programs w. DA (ours) & \textbf{0.0035} & \textbf{0.0043} \\
    \bottomrule
    \end{tabular}
    \vskip -0.1in
    \label{tab:performance}
\end{table}

\section{Conclusion}
In this work, we present an analytical differentiable renderer for more accurate rendering and more efficient adaptation of the neural program generator that requires no training.
We show that, with preliminary results, it has prospects of becoming a useful member of the \emph{shape program} toolkit.
While our executor is presented as an alternative to the neural executor, one may also wish to experiment with the utility of using a hybrid of them which we leave for future work.
We hope this is also a step towards learning shape program libraries, as, in contrast to neural program interpreters, our renderer can be flexibly modified to precisely interpret new program statements without any training.
In future works, we hope to realized library learning of shape programs, and experiment with other differentiable reconstruction representations and learning objectives.

\section{Acknowledgment}
We thank our advisor Bob Fisher for his constant support, insightful discussion throughout this project, and helpful comments on early manual scripts.

\bibliographystyle{abbrvnat}
\bibliography{main}

\newpage
\appendix

\section{Distance Field of Cylinders}\label{app:cyl_df}
\begin{figure}
    \centering
    \includegraphics[width=0.3\textwidth]{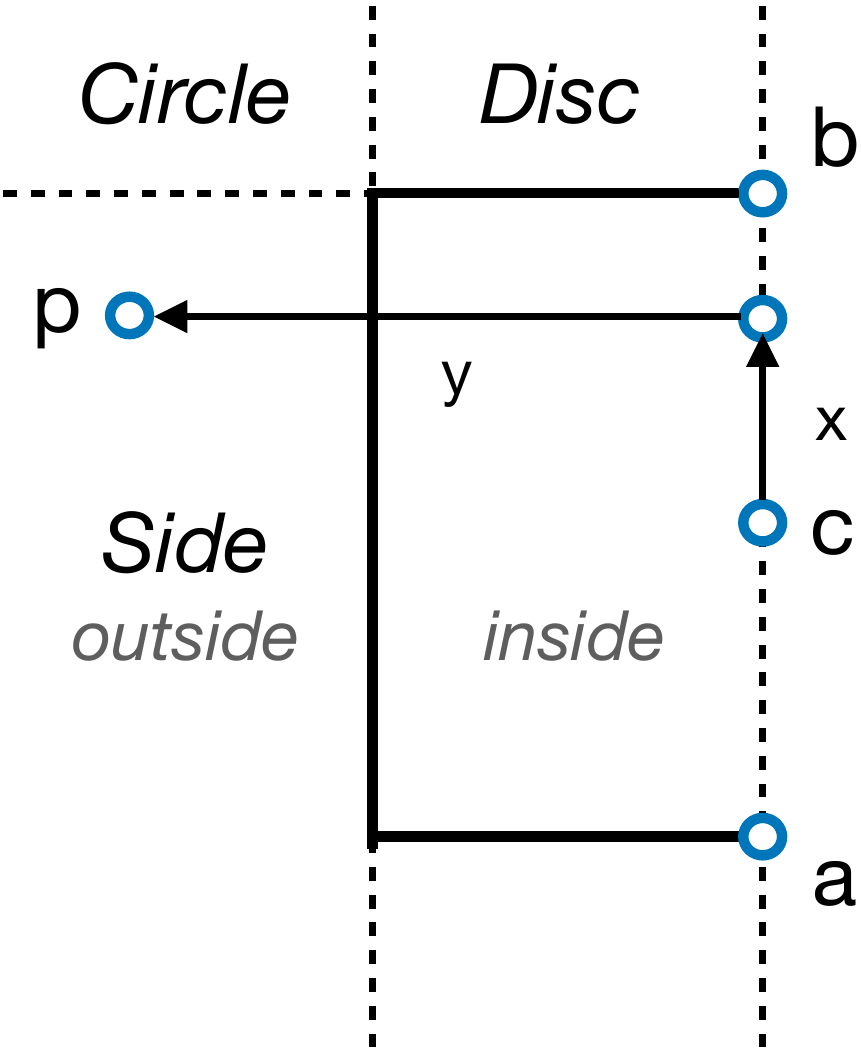}
    \caption{Depicting when a point $p$ is on the side (outside) of a cylinder.}
    \label{fig:cyl_df}
\end{figure}

\cref{alg:cyldf} describes the algorithm for evaluating the distance field of cylinders.
\cref{fig:cyl_df} demonstrates one of the four possible cases; when a point $p$ is on the side and within the height of the cylinder).

\begin{algorithm}[tb]
   \caption{Distance Field of a Cylinder}
   \label{alg:cyldf}
\begin{algorithmic}
   \STATE {\bfseries Input:} point $p$, cylinder\_shape $s_x, s_y$
   \STATE {\bfseries Return:} squared\_distance $d$
   \STATE Initialize $c=[0,0,0], a=[s_x,0,0], b=[-s_x,0,0]$ \COMMENT{c denotes the origin/center of the cylinder, a,b the center of the top/bottom disc of the cylinder}
   \STATE Initialize $u =(b-a)/||b-a||$ \COMMENT{unit vector in the direction of b-a}
   \STATE $x = (c - p)\cdot u$ \COMMENT{projection onto u} 
   \STATE $n^2 = (c-p)^2$ \COMMENT{squared distance from c to p}
   \STATE $y^2 = n^2 - x^2$ \COMMENT{squared distance from p to u-axis}
   \IF{$|x| =< s_x/2$} 
   \STATE \COMMENT{if p is on the side of the cylinder}
   
   \IF{$y^2 =< r^2$} 
   \STATE \COMMENT{if p is inside of the cylinder}
   \STATE \bfseries{return} $d = 0$
   \ELSE 
   \STATE \COMMENT{p is outside of the cylinder}
   \STATE \bfseries{return} $d = \sqrt{(y-r)^2}$
   \ENDIF
   \ENDIF
   
   \IF{$|x|>s_x/2$} 
   \STATE \COMMENT{otherwise}
   \IF{$y^2 < s_x^2$}
   \STATE \COMMENT{if p is in the Disc region}
   \STATE \bfseries{return} $d = \sqrt{(|x|-s_x)^2}$
   \ELSE
   \STATE \COMMENT{else its in the Circle region}
   \STATE \bfseries{return} $d = \sqrt{(y-r)^2+(|x|-s_x/2)^2}$
   \ENDIF
   \ENDIF
\end{algorithmic}
\end{algorithm}

\appendix
\end{document}